\documentclass[11pt]{article}
\usepackage{acl2016}
\usepackage{times}
\usepackage{url}
\usepackage{latexsym}
\usepackage{amsmath,amssymb,amsthm}
\usepackage{graphics}
\usepackage{graphicx}
\usepackage{placeins}
\usepackage{booktabs}
\usepackage{numprint}
\usepackage{float}
\usepackage{fullpage}
\usepackage{tikz}
\usetikzlibrary{arrows,shapes,positioning,backgrounds}

\aclfinalcopy

\newcommand{\set}[1]{\{#1\}}
\newcommand*{\Scale}[2][4]{\scalebox{#1}{$#2$}}

\theoremstyle{definition}

\title{On the Linearity of Semantic Change: 
  Investigating Meaning Variation via Dynamic Graph Models 
}

  \author{Steffen Eger \\ Ubiquitous Knowledge Processing
    Lab\\ Department of Computer Science \\ Technische Universit\"at
    Darmstadt \And Alexander Mehler \\ Text Technology Lab
    \\ Department of Computer Science \\ Goethe-Universit\"at Frankfurt
    am Main}

\date{}

\begin{document}
\maketitle

\begin{abstract}
We consider two graph models of semantic change. The first is a
time-series model that relates embedding vectors from one time period
to embedding vectors of previous time periods. In the second, we
construct one graph for each word: nodes in this graph correspond
to time points and edge weights to the similarity of the word's
meaning across two time points. We apply our two models to 
corpora across three different languages. We find that
semantic change is \emph{linear} in two senses. Firstly, today's
embedding vectors ($=$ meaning) of words can be derived as linear
combinations of embedding vectors of their neighbors in previous time
periods. Secondly, self-similarity of words decays linearly in
time. We consider both findings as new laws/hypotheses of semantic
change. 
\end{abstract}

\section{Introduction}\label{sec:introduction}
Meaning is not uniform, neither across space, nor across time. Across
space, different languages tend to exhibit different polysemous
associations for corresponding terms
\cite{Eger:Schenk:Mehler:2015,Kulkarni:2015b}. 
Across time, several well-known
examples 
of meaning change in English have been documented. For example, the
word \emph{gay}'s meaning has shifted, during the 1970s, from an 
adjectival meaning of 
\emph{cheerful} at the beginning of the 20$^{\text{th}}$ century to its present
meaning of \emph{homosexual} \cite{Kulkarni:2015}. Similarly,
technological progress has led to semantic broadening of terms
such as \emph{transmission}, \emph{mouse}, or \emph{apple}. 

In this work, we consider two graph models of semantic
change.
Our \textbf{first} model is a \emph{dynamic} model in that the underlying paradigm
is a (time-)series of graphs. Each node in the series of graphs
corresponds to one word, associated with which is a semantic {embedding}
vector. We then ask how the embedding vectors in one time period (graph) can
be predicted from the embedding vectors of neighbor words in previous
time periods. In particular, we postulate that there is a linear
functional 
relationship that couples a word's today's meaning with its neighbor's
meanings in the past. When estimating the coefficients of this
model, we find that the linear form appears indeed very plausible. This
functional form then allows us to address further questions,
such as negative relationships between words --- which indicate
semantic differentiation over time --- as well as projections into the
future. We call our \textbf{second} graph model \emph{time-indexed self-similarity
graphs}. In these graphs, each node corresponds to a time point and the link
between two time points indicates the semantic similarity of a
specific word across the two time points under consideration. The analysis of
these graphs reveals that most words obey a law of linear semantic
`decay': semantic self-similarity decreases linearly over time. 

In our work, we capture semantics by means of word embeddings
derived from context-predicting neural network architectures, which
have become the state-of-the-art in distributional semantics
modeling \cite{Baroni:2014}. 
Our approach and results are partly independent of this representation, 
however, in
that we 
take a structuralist approach: we derive new, `second-order
embeddings' 
by
modeling the meaning of words by means of their semantic similarity relations to
all other words in the
vocabulary \cite{Saussure:1916,Rieger:2003}. 
Thus, future research may in principle substitute the deep-learning
architectures for semantics considered here
by any other method capable of producing semantic similarity values between
lexical units. 

This work is structured as follows. In \S \ref{sec:related}, we
discuss related work. In \S \ref{sec:model1} and \ref{sec:model2},
respectively, we formally introduce the two graph
models outlined. In \S \ref{sec:experiments}, we detail our
experiments and in \S \ref{sec:conclusion}, we conclude.

\section{Related work}\label{sec:related}
Broadly speaking, one can distinguish two recent NLP approaches to meaning change
analysis. 
On the one hand, \emph{coarse-grained} trend analyses 
compare the semantics of a
word in one time period with the meaning of the word in the preceding time
period \cite{Jatowt:2014,Kulkarni:2015}. Such coarse-grained models,
by themselves, do
not specify \emph{in which respects} a word has changed (e.g.,
semantic broadening or narrowing), but 
just aim at capturing whether meaning change has occurred. In
contrast, more fine-grained analyses typically sense-label word
occurrences in corpora and then investigate changes in the
corresponding meaning 
distributions \cite{Rohrdantz:2011,Mitra:2014,Poelitz:2015,Zhang:2015}. Sense-labeling may be achieved by
clustering of the context vectors of words 
\cite{Huang:2012,Chen:2014,Neelakantan:2014} or by applying
LDA-based techniques where word contexts take the roles of documents
and word senses take the roles of topics \cite{Rohrdantz:2011,Lau:2012}. 
Finally, there are studies that test particular meaning change
hypotheses such as whether similar words tend to diverge in meaning
over time (according to the `law of differentiation') \cite{Xu:2015}
and papers that intend to detect corresponding terms across time
(words with similar meanings/roles in two time periods but potentially
different lexical 
forms) \cite{Zhang:2015}.

\section{Graph models}\label{sec:models}
Let $V=\set{w_1,\ldots,w_{|V|}}$ be the common vocabulary 
(intersection) of 
all words in all time periods $t\in\mathcal{T}$.
Here, $\mathcal{T}$ is a set of time indices. 
Denote an embedding of a word $w_i$ at time
period $t$ as $\mathbf{w}_{i}(t)\in\mathbb{R}^d$. 
Since embeddings
$\mathbf{w}_{i}(s),\mathbf{w}_{i}(t)$ for two different time periods $s,t$ are
generally not comparable, as they may lie in different coordinate
systems, we consider the vectors $\tilde{\mathbf{{w}}}_{i}(t)=$
\begin{align}\label{eq:repr}
  \Scale[0.90]{
   \left(
  \mathsf{sim}(\mathbf{w}_{i}(t),\mathbf{w}_{1}(t)),\ldots,
  \mathsf{sim}(\mathbf{w}_{i}(t),\mathbf{w}_{|V|}(t))
  \right)},
\end{align}
each of which lies in $\mathbb{R}^{|V|}$ and where $\mathsf{sim}$ is a
similarity function such as the cosine. 
We note
that our structuralist 
definition of $\tilde{\mathbf{w}}_{i}(t)$ is not unproblematic, since the
vectors $\mathbf{w}_{1}(t),\ldots,\mathbf{w}_{|V|}(t)$ tend to
be different across $t$, by our very postulate, so that there is
non-identity of these `reference
points' over time. 
However, as we may
assume that the meanings of at least a few words are stable over time, we
strongly expect the vectors $\tilde{\mathbf{w}}_{i}(t)$ 
to be suitable for our task of analysis of meaning changes.\footnote{An
  alternative to our second-order embeddings is to project vectors
  from different time periods in a common 
  space \cite{Mikolov:2013b,Faruqui:2014}, which requires to find
  corresponding terms across time. 
  Further, one could also consider a `core' vocabulary
  of semantically stable words, e.g., in the spirit
  of \newcite{Swadesh:1952}, instead of using all vocabulary words 
  as reference.} For the 
remainder of this work, for convenience, we do not distinguish, in terms
of notation, 
between 
$\mathbf{w}_{i}(t)$ and  
$\tilde{\mathbf{w}}_{i}(t)$.
\subsection{A linear model of semantic change}\label{sec:model1}
We 
postulate, and subsequently test, the following 
model of meaning dynamics 
which describes meaning change over time for words $w_i$: 
\begin{align}\label{eq:average}
  \mathbf{w}_i(t) = \sum_{n=1}^p\sum_{w_j\in V\cap N(w_i)}
  \alpha_{w_j}^n\mathbf{w}_j({t-n})
\end{align}
where $\alpha_{w_j}^n\in\mathbb{R}$, for $n=1,\ldots,p$, are coefficients
of meaning vectors 
$\mathbf{w}_j({t-n})$ and $p\ge 1$ is the \emph{order} of the model. 
The set $N(w_i)\subseteq V$ denotes a set of `neighbors' of word
$w_i$.\footnote{We also constrain the vectors $\mathbf{w}_i(t)$, for all
  $w_i\in V$, to contain
  non-zero entries only for words in $N(w_i)$.}
This model says
that the meaning of a word $w_i$ at some time $t$ is determined by
reference to the meanings of its `neighbors' in previous time periods,
and that the underlying functional relationship is \emph{linear}. 

We remark that the model described by Eq.~\eqref{eq:average} is 
a time-series model, and, in particular, a vector-autoregressive (VAR)
model with special structure. The model may also be seen in the
socio-economic context of so-called ``opinion dynamic
models'' \cite{Golub:2010,Acemoglu:2011,Eger:2016}. There it is
assumed that agents are situated in network structures and
continuously update their opinions/beliefs/actions according to their
ties with other agents. Model \eqref{eq:average} substitutes
multi-dimensional embedding vectors for 
one-dimensional opinions.  
\subsection{Time-indexed
  self-similarity 
  graphs}\label{sec:model2} 
We track meaning change by considering a fully connected graph
$G(w)$ for each word $w$ in $V$. The nodes of $G(w)$ are the time 
indices $\mathcal{T}$, and there is an undirected link between any two
$s,t\in \mathcal{T}$ whose weight is given by 
  $\mathsf{sim}(\mathbf{w}({s}),\mathbf{w}(t))$.
We call the graphs $G(w)$ \emph{time-indexed self-similarity (TISS) graphs}
because 
they indicate the (semantic) similarity of a given word with itself
across 
different time periods. Particular interest may lie in \emph{weak
  links} in these graphs as they indicate low similarity between two
different time periods, i.e., semantic change across time.

\section{Experiments}\label{sec:experiments}
\textbf{Data} 
As corpus for English, we use the Corpus of Historical American (COHA).\footnote{http://corpus.byu.edu/coha/.} This 
covers texts from the time period 1810 to 2000. We extract two 
slices: the years 1900-2000 and 1810-2000. 
For both 
slices, each time period $t$ is one decade, e.g.,
$\mathcal{T}=\set{1810,1820,1830,\ldots}$.\footnote{Each time period
contains texts that were written in that decade.}
For each slice, we only keep
words associated to the word classes nouns, adjectives, and verbs. For
computational and estimation purposes, we also only consider words
that occur at least 100 times in each time period.
To induce word embeddings $\mathbf{w}\in\mathbb{R}^d$ for each word $w\in
V$,  
we use
word2vec \cite{Mikolov:2013} with default parametrizations. We do so
for each time period 
$t\in\mathcal{T}$ 
independently. We then use these embeddings to derive the new
embeddings as in 
Eq.~\eqref{eq:repr}.  
Throughout, we use 
cosine similarity as $\mathsf{sim}$ measure. For German, we consider a
proprietary dataset of the German
newspaper \emph{SZ}\footnote{http://www.sueddeutsche.de/} for which 
$\mathcal{T}=\set{1994,1995,\ldots,2003}$. We lemmatize and POS tag
the data and likewise only consider nouns, verbs and adjectives,
making the same frequency constraints as in English. Finally, we use
the PL \cite{Migne:1855} as data set for Latin. Here,
$\mathcal{T}=\set{300,400,\ldots,1300}$. We use the same
preprocessing, frequency, and word class constraints as for English
and German.

Throughout, our datasets are well-balanced in terms of size. For
example, the 
English COHA datasets contain about 24M-30M tokens for each decade
from 1900 to 2000, where the decades 1990 and
2000 contain slighly more data than the earlier decades. The pre-1900
decades contain 18-24M tokens, with only the decades 1810 and 1820
containing very little data (1M and 7M tokens, respectively). The
corpora are also balanced by genre.   

\subsection{TISS graphs}
\begin{figure}[!htb]
\begin{center}
\input{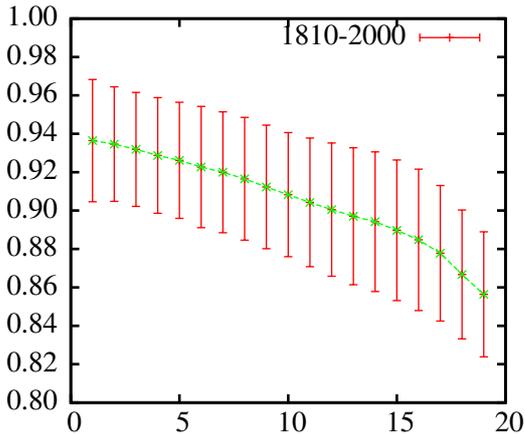}
\end{center}
\caption{{$D_{t_0}$ ($y$-axis) as a function of $t_0$
($x$-axis), values of $D_{t_0}$ (in green) and error-bars.}}
\label{fig:years}
\end{figure}

We start with investigating the TISS graphs. 
Let $D_{t_0}$ represent how semantically
similar a word is across two time periods, on average, when the
distance between time periods is $t_0$: 
$D_{t_0}=\frac{1}{C}\sum_{w\in V}\sum_{|s-t|=t_0}\mathsf{sim}({\mathbf{w}(s),\mathbf{w}(t)})$, where 
$C=|V|\cdot|\set{(s,t)\,|\,|s-t|=t_0}|$ is 
a normalizer. 
Figure \ref{fig:years} plots the values $D_{t_0}$ for the time slice
from 1810 to 2000, for the English data.  
We notice a clear trend: self-similarity
of a word tends to (almost perfectly) linearly decrease with time distance.  
In fact, Table \ref{table:change} below
indicates 
that this trend holds
across all our corpora, i.e., for different time scales and different
languages: the linear `decay' model fits the $D_{t_0}$ curves very
well, with adjusted $R^2$ values substantially above 90\% and consistently
and significantly negative coefficients. We believe that this finding
may be considered a new statistical law of semantic change. 

\begin{table}[!htb]
  \begin{center}
    {\small 
  \begin{tabular}{|p{0.55cm}p{1cm}p{1cm}p{1.38cm}p{0.7cm}c|}\hline
    Corpus & Lang. & Time interval & Years & Coeff. & $R^2$\\ \hline
    COHA & English & Decade & 1900-2000 & $-$0.425 & 98.63 \\
         &         &        & 1810-2000 & $-$0.405 & 96.03 \\
    SZ   & German  & Year   & 1994-2003 & $-$0.678 & 98.64 \\
    PL   & Latin   & Century & 400-1300 & $-$0.228 & 92.28 \\ \hline
  \end{tabular}
  \caption{Coefficients (\%) in regression of $D_{t_0}$ on $t_0$, and
    adjusted $R^2$ values (\%).}
  \label{table:change}
  }
  \end{center}
\end{table}

\begin{figure}[!htb]
  \begin{center}
  \input{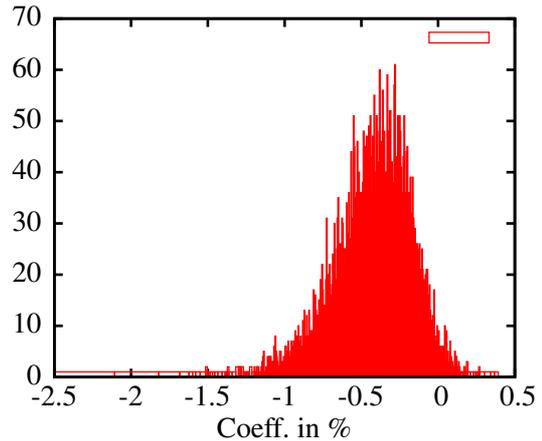}
  \input{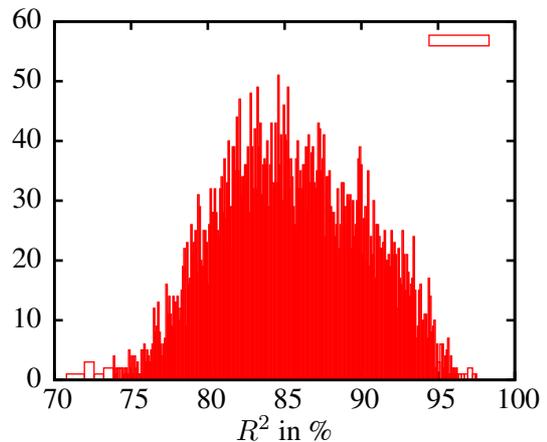}
  \end{center}
  \caption{Distribution of Coefficients $\alpha$ (top) and $R^2$
  values (bottom) in
  regression of values $D_{t_0}({w})$ on $t_0$. 
  The plots are histograms: $y$-axes are 
  frequencies.
  }
  \label{fig:histograms}
\end{figure}

The values $D_{t_0}$ as a function of $t_0$ are 
averages over all words. Thus, it might be possible that the
average word's meaning decays linearly in time, while the semantic
behavior, over time, of 
a large fraction of words follows different trends. To investigate
this, we consider the distribution of
$D_{t_0}(w)=\frac{1}{C'}\sum_{|s-t|=t_0}\mathsf{sim}(\mathbf{w}(s),\mathbf{w}(t))$
over fixed words $w$. Here
$C'=|\set{(s,t)\,|\, |s-t|=t_0}|$.
We consider the regression models
\begin{align*}
        D_{t_0}(w) = \alpha\cdot t_0+\text{const.}
\end{align*}
for each word $w$ independently and assess the distribution of
coefficients $\alpha$ as well as the 
goodness-of-fit values. Figure \ref{fig:histograms} shows ---
exemplarily for the English 1900-2000 COHA data --- that the
coefficients $\alpha$ are negative for almost all words. In fact, the
distribution is left-skewed with a mean of around $-0.4\%$. Moreover,
the linear model is always a good to very good fit of the data in that 
$R^2$ values are centered around 85\% and rarely fall below
75\%. We find similar patterns for all other datasets considered
here. 
This shows that not only the average word's meaning decays linearly, but almost
all words'  (whose frequency mass exceeds a particular
threshold) semantics behaves this way. 

Next, 
we use our TISS graphs for the task of finding
words that have undergone meaning change. To this end, 
we sort the graphs 
$G(w)$ by the ratios
$R_{G(w)}=\frac{\text{maxlink}}{\text{minlink}}$, where
$\text{maxlink}$ denotes 
maximal weight of a link in graph $G(w)$ and $\text{minlink}$ is the
minimal weight of a link in graph $G(w)$. 
We note that weak links may indicate semantic change, but the
stated ratio requires that `weakness' is seen relative to the
strongest semantic links in the TISS graphs. 
Table \ref{table:1810-1900} presents selected words that have 
highest values $R_{G(w)}$.\footnote{The top ten
words with the lowest values $R_{G(w)}$ are \emph{one, write, have,
who, come, only, even, know, hat, fact}. }  
\begin{table}[!htb]
  \begin{center}
  \begin{tabular}{|l|l|l|}\hline
    bush (1), web (2), alan (3), implement (4)\\ 
    jeff (5), gay (6), program (7), film (8), \\
    focus (9), terrific (16), axis (36)\\ 
    \hline
  \end{tabular}
  \caption{Selected words with highest values $R_{G(w)}$ in COHA
    for the 
    time period 1900-2000. In brackets are the ranks of words, i.e., \emph{bush} has the highest value $R_{G(w)}$, \emph{web} the 2nd highest, etc.}
  \label{table:1810-1900}
  \end{center}
\end{table}
We omit a fine-grained semantic change analysis, which could be
conducted via the methods outlined in
\S\ref{sec:related}, but notice a few cases. 
`Terrific' has a large semantic discrepancy between the 1900s 
and the 1970s, when the word probably (had) changed from a negative to a
more positive meaning. The largest discrepancy for `web' is between
the 1940s and the 2000s, when it probably came to be massively used in
the context of the Internet. The high $R_{G(w)}$ value for $w=$ `axis' 
derives from comparing its use in the 1900s with its use in the 1940s,
when it probably came to be used in the context of Nazi Germany and
its allies. We notice that the presented method can account for gradual,
accumulating change, which is not possible for models that compare two
succeeding time points such as the model of \newcite{Kulkarni:2015}. 

\subsection{Meaning dynamics network models}
Finally, we estimate meaning dynamics models as in
Eq.\ \eqref{eq:average}, i.e., we estimate the coefficients
$\alpha_{w_j}^n$ from our data sources. We let the neighbors $N(w)$ of
a word $w$ as in 
Eq.\ \eqref{eq:average} be the
union (w.r.t.\ $t$) over sets $N_t(w;n)$ 
denoting the $n\ge 1$
semantically 
most similar words (estimated by cosine similarity on the original
word2vec vectors) of word $w$ in 
time period $t\in\mathcal{T}$.\footnote{We exclude word $w$ from
$N_t(w;n)$. We found that including $w$ did not improve performance
results.} In Table \ref{table:r2}, we indicate 
two measures: adjusted $R^2$, which indicates the goodness-of-fit of a
model, and prediction error. By prediction error, we measure the
average Euclidean distance between the true semantic vector
of a word in 
the \emph{final} time period $t_N$ vs.\ the predicted semantic vector,
via the linear model in Eq.~\eqref{eq:average}, 
estimated on the data excluding the final period. The indicated
prediction error is the average over all words. We note the following:
$R^2$ values are high (typically above 95\%), indicating that the
linear semantic change model we have suggested fits the data well. 
Moreover, $R^2$ values slightly increase between order $p=1$ and $p=2$;
however, for prediction error this trend is reversed.\footnote{We
experimented with 
orders $p\ge 3$, but found them to be inadequate. Typically,
coefficients for lagged-$3$ variables are either zero or model
predictions are way off, possibly indicating multi-collinearity.} 
We also include a strong baseline that claims that word
meanings do not change in the final period $t_N$ but are the same as
in $t_{N-1}$. We note that the order $p=1$ model consistently improves
upon this baseline, by as much as 18\%, depending upon parameter
settings.   
\begin{table}[!htb]
\begin{center}
{\small 
\begin{tabular}{ccccc}
  $n$ & $p$ & Adjusted-$R^2$ & Pred.\ Error & Baseline\\ \hline
  5 & 1    & 95.68$\pm$ 2.80 & 0.402$\pm$.234 & 0.447$\pm$.232\\
     & 2   & 96.13$\pm$ 1.83 & 0.549$\pm$.333 \\
  10 & 1   & 95.24$\pm$ 2.78 & 0.378$\pm$.169 & 0.445$\pm$.187\\
     & 2   & 95.75$\pm$ 2.67 & 0.515$\pm$.247 \\
  20 & 1   & 94.72$\pm$ 2.85 & 0.362$\pm$.127 & 0.442$\pm$.156\\
     & 2   & 95.27$\pm$ 2.74 & 0.493$\pm$.190
\end{tabular}
}
\caption{English data, 1900-2000. $R^2$ and prediction error in \%.}
\label{table:r2}
\end{center}
\end{table}

\textbf{Negative relationships}
Another very interesting aspect of the model in Eq.~\eqref{eq:average} is that it
allows for detecting words $w_j$ whose embeddings have
negative coefficients $\alpha_{w_j}$ for a target word $w_i$ (we consider
$p=1$ in the remainder). Such negative coefficients may be seen as
instantiations of the `law of differentiation': the two words' meanings
move, over time, in opposite directions in semantic space. We find
significantly negative relationships between the following words,
among others: 
summit $\leftrightarrow$ foot, boy $\leftrightarrow$ woman, vow
$\leftrightarrow$ belief, negro $\leftrightarrow$ black. 
Instead of a detailed analysis, we mention that the Wikipedia
entry for the last pair indicates that the meanings of `negro' and
`black' switched roles between the early and late 20$^{\text{th}}$ 
century. While `negro' was once the ``neutral'' term for the colored
population in the US, it acquired negative connotations after the
1960s; and vice versa for `black'.

\section{Concluding remarks}\label{sec:conclusion}
We suggested two novel models of semantic change. 
First, TISS graphs allow for detecting gradual, non-consecutive meaning
change. 
They enable to detect statistical trends of a possibly general nature. 
Second, our time-series models allow for investigating negative trends
in meaning change (`law of differentiation') as well as 
forecasting into the future, which we leave for future work.   
Both models hint at a linear behavior of semantic change, which
deserves further 
investigation.
We note that this linearity 
concerns the \emph{core} vocabulary of languages (in
our case, words that occurred at least 100 times in each time period), and, in
the case of TISS graphs, is an \emph{average} result; particular words
may have drastic, non-linear meaning changes across time (e.g., proper
names referring to entirely different entities). 
However, our analysis also finds that 
most core words' meanings
decay linearly in time.

\FloatBarrier

\bibliographystyle{acl2016}
\bibliography{lit}

\end{document}